\documentclass[conference]{IEEEtran}
\IEEEoverridecommandlockouts
\usepackage[a-1b]{pdfx}
\usepackage{balance}
\usepackage{hyperref} 
\usepackage{booktabs} 
\usepackage{xcolor}    
\usepackage{colortbl} 
\usepackage{subcaption}
\usepackage{cite}
\usepackage{amsmath,amssymb,amsfonts}
\usepackage{algorithmic}
\usepackage{graphicx}
\usepackage{textcomp}
\def\BibTeX{{\rm B\kern-.05em{\sc i\kern-.025em b}\kern-.08em
    T\kern-.1667em\lower.7ex\hbox{E}\kern-.125emX}}
\begin{document}

\title{Synthetic Aircraft Trajectory Generation Using Time-Based VQ-VAE
\thanks{This paper was presented at the 25th Integrated Communications, Navigation and Surveillance 
Conference (ICNS 2025), April 8--10, 2025, Brussels, Belgium.}}

\author{
    \IEEEauthorblockN{Abdulmajid Murad$^{1}$, Massimiliano Ruocco$^{1,2}$}
    \IEEEauthorblockA{
        \parbox{0.5\textwidth}{
            \centering
            $^1$Department of Software Engineering, Safety and Security \\ 
            SINTEF Digital \\
            Trondheim, Norway \\
            \{abdulmajid.murad, massimiliano.ruocco\}@sintef.no
        }
        \hspace{0.05\textwidth}
        \parbox{0.4\textwidth}{
            \centering
            $^2$Department of Computer Science \\
            Norwegian University of Science and Technology \\
            Trondheim, Norway \\
            massimiliano.ruocco@ntnu.no
        }
    }
}

\maketitle

\begin{abstract}
In modern air traffic management, generating synthetic flight trajectories has emerged as a promising solution for addressing data scarcity, protecting sensitive information, and supporting large-scale analyses.  In this paper, we propose a novel method for trajectory synthesis by adapting the Time-Based Vector Quantized Variational Autoencoder (TimeVQVAE). Our approach leverages time-frequency domain processing, vector quantization, and transformer-based priors to capture both global and local dynamics in flight data. By discretizing the latent space and integrating transformer priors, the model learns long-range spatiotemporal dependencies and preserves coherence across entire flight paths.
We evaluate the adapted TimeVQVAE using an extensive suite of quality, statistical, and distributional metrics, as well as a \emph{flyability} assessment conducted in an open-source air traffic simulator.
Results indicate that TimeVQVAE outperforms a temporal convolutional VAE baseline, generating synthetic trajectories that mirror real flight data in terms of spatial accuracy, temporal consistency, and statistical properties. 
Furthermore, the simulator-based assessment shows that most generated trajectories maintain operational feasibility, although occasional outliers underscore the potential need for additional domain-specific constraints. Overall, our findings underscore the importance of multi-scale representation learning for capturing complex flight behaviors and demonstrate the promise of TimeVQVAE in producing representative synthetic trajectories for downstream tasks such as model training, airspace design, and air traffic forecasting.

\end{abstract}

\begin{IEEEkeywords}
    Aircraft trajectory, synthetic data generation, multivariate time series, machine learning in ATM.
\end{IEEEkeywords}

\section{Introduction}

As air traffic continues to grow in complexity and volume, there is an increasing need for advanced tools to model and analyze flight operations. In particular, data-driven research in air traffic management (ATM) often requires large amounts of historical flight data. However, due to privacy concerns, data accessibility restrictions, and the relative scarcity of certain flight scenarios, obtaining sufficiently comprehensive datasets remains a challenge. Synthetic trajectory generation has emerged as a promising approach to address these issues, as it can produce additional flight data for augmenting machine learning models and conducting broader airspace analyses without disclosing sensitive information.

Existing work on synthetic trajectory generation spans a variety of approaches, ranging from physics-based or model-driven simulations \cite{delahaye2014mathematical} to purely data-driven methods \cite{krauth2023deep, ezzahed2022bringing, krauth2024advanced} that learn flight patterns directly from historical data. While model-driven approaches can provide interpretable results grounded in physical equations, their applicability is limited by the accuracy of the underlying assumptions and external factors such as weather conditions. On the other hand, data-driven methods leverage deep learning architectures to automatically extract complex spatiotemporal structures from real flight trajectories. Yet, many of these approaches struggle to capture long-range dependencies and maintain consistency throughout entire flight paths.

In this paper, we propose the adaptation of the Time-Based Vector Quantized Variational Autoencoder (TimeVQVAE) \cite{lee_vector_2023} for the task of synthetic aircraft trajectory generation. This model processes flight data in the time-frequency domain, allowing it to represent both global trends (e.g., overall flight paths) and local variations (e.g., altitude adjustments and minor maneuvers). The use of transformer-based priors enable the model to capture the temporal dependencies inherent in flight paths, ensuring that the generated trajectories remain coherent over long distances and throughout various phases of flight. Additionally, using MaskGIT \cite{chang_maskgit_2022} with iterative decoding accelerates the generation process and fosters the generation of diverse flight scenarios. This combination of techniques allows the adapted TimeVQVAE model to effectively balance fidelity and diversity, offering a scalable solution for synthetic trajectory generation.

We evaluate our approach using various assessment strategies, including quality and statistical metrics commonly used in time series generation. Additionally, we incorporate a \emph{flyability} assessment using an open-source air traffic simulator to verify whether the generated trajectories adhere to basic physics and operational constraints. Our results show that the TimeVQVAE-based approach outperforms other methods, such as the temporal convolutional VAE (TCVAE)-based approach \cite{krauth2023deep}, in both trajectory fidelity and diversity.

Our main contributions include:
\begin{itemize}
    \item Adapting the TimeVQVAE model for the domain of aircraft trajectory generation.
    \item Proposing a robust evaluation framework for assessing synthetic trajectory quality.
    \item Conducting a comparative analysis with a TCVAE-based model.
    \item Providing an open-source implementation of our models and evaluation pipeline.\footnote{Code available at: \href{https://github.com/SynthAIr/T-VQ-VAE-TrajGen}{https://github.com/SynthAIr/T-VQ-VAE-TrajGen}}

\end{itemize}

By offering a data-driven approach for synthetic trajectory generation, this work aims to facilitate more effective machine learning pipelines and airspace studies, while acknowledging the complexities and limitations that come with modeling real-world air traffic phenomena.

\section{Related Work}

Synthetic aircraft trajectory generation approaches in the literature can be broadly categorized into model-driven and data-driven methods, each with distinct strengths and limitations.

Model-driven approaches, such as those described by Delahaye et al. \cite{delahaye2014mathematical}, rely on flight mechanics principles to generate trajectories optimized for specific objectives like fuel efficiency or time minimization. These methods require detailed knowledge of aircraft performance characteristics and environmental factors, such as weather and traffic conditions. While model-driven approaches offer a structured and deterministic framework for simulating flight paths, they often fail to capture the full variability and complexity of real-world air traffic operations. As a result, they are less flexible in modeling the diverse and irregular flight patterns observed in practice, particularly in dynamic environments where conditions frequently change.

In contrast, data-driven approaches have gained significant attention in recent years due to their ability to leverage historical flight data to generate large volumes of feasible and statistically representative flight paths. Data-driven methods can be further divided into two subcategories: statistical models and deep learning-based generative models. Early work in data-driven trajectory generation primarily focused on statistical methods for density estimation. For instance, Krauth et al. \cite{krauth2021synthetic} employed dimensionality reduction and statistical copulas to model two-dimensional flows, generating synthetic trajectories that mimic real flight behaviors in terminal maneuvering areas. Similarly, Murça et al. \cite{murca2020data} applied Gaussian Mixture Models (GMMs) to predict future aircraft positions based on partial trajectory observations, while Jacquemart et al. \cite{jacquemart2013conflict} used Monte Carlo simulations and dynamic importance splitting to model aircraft trajectories as stochastic processes, estimating conflict probabilities in uncontrolled airspace. Although these statistical approaches provide valuable insights into flight behavior and are effective for modeling basic trajectory patterns, they often struggle to capture the full procedural fidelity and operational complexities of modern air traffic management systems.

The advent of deep learning has introduced more advanced data-driven generative models for synthetic trajectory generation.These models harness the representational power of neural networks to learn complex, non-linear relationships in-flight data, enabling them to generate new and more realistic synthetic trajectories. Variational Autoencoders (VAEs) have emerged as a key tool in this domain. For instance, Krauth et al. \cite{krauth2023deep}, Ezzahed et al. \cite{ezzahed2022bringing}, and Krauth et al. \cite{krauth2024advanced} demonstrated the use of VAEs in modeling aircraft trajectories within terminal maneuvering areas. These models effectively capture the underlying structure of flight data and generate statistically consistent synthetic trajectories that align closely with observed real-world data. Despite the advances in deep learning, many of these generative models are still focused on specific phases of the flight path, such as terminal areas, and often lack the capacity to generalize across the entire flight trajectory.

Building on these data-driven techniques, our work investigates a Time-Based Vector Quantized VAE (TimeVQVAE) to address some of the challenges in synthetic trajectory generation. Specifically, we target full-flight trajectories from takeoff to landing, aiming to capture both high-level temporal phases and finer local details. TimeVQVAE, despite being a recent development, has shown strong performance in benchmarks for time series generation tasks \cite{ang_tsgbench_2023} and has been adapted for other applications, such as time series anomaly detection \cite{lee_explainable_2024}. By processing the data in the time-frequency domain and utilizing transformer-based priors, the model can learn long-range dependencies and produce synthetic trajectories that exhibit some of the spatial and temporal characteristics observed in real data.

\section{Methodology}

\subsection{Problem Definition}
The goal of this study is to develop a generative model that can produce synthetic aircraft trajectories. We formally define this task as follows:

Given a dataset
\[
D = \{\tau_1, \tau_2, \dots, \tau_n\}
\]
of \(n\) real aircraft trajectories, where each trajectory \(\tau\) is a sequence of \(m\) time-stamped points
\[
\tau = \bigl\{(x_1, y_1, z_1, t_1), \dots, (x_m, y_m, z_m, t_m)\bigr\},
\]
and \((x, y, z)\) are the 3D spatial coordinates at time \(t\). Our aim is to learn a generative model \(G\) that produces synthetic trajectories 
\[
\tau' \sim G(z),
\]
given a latent representation \(z\). In doing so, we seek to:
\begin{enumerate}
    \item Ensure that the overall distribution of the synthetic set \(P(\tau')\) approximates the real-world data distribution \(P(\tau)\).
    \item Preserve temporal and spatial coherence throughout each trajectory, mirroring essential trends observed in actual flight paths.
    \item Adhere to fundamental physical constraints and prevalent operational norms in air traffic.
    \item Provide sufficient diversity to reflect key variations found in actual flight data.
\end{enumerate}

Beyond these core objectives, we also explore \emph{class-conditional} generation, where the model can be guided by a specified attribute or class label \(c\). Formally, this amounts to generating 
\[
\tau' \sim G(z, c),
\]
with the intent of producing trajectories that conform to designated categories (e.g., particular routes or flight profiles). This conditional capability is particularly useful for controlled and targeted trajectory generation in various ATM scenarios and analyses.

\subsection{Data Acquisition and Preprocessing}

We draw from two main data sources—OpenSky Network \cite{schafer2014bringing} and EUROCONTROL R\&D Archive \cite{noauthor_aviation_2023}—to assemble end-to-end flight trajectories. These datasets offer substantial coverage and diversity of flight paths.

\textbf{OpenSky Network.} As our primary source, OpenSky provides high-resolution historical flight data. To create a consistent working dataset, we perform several processing steps:
\begin{itemize}
    \item \emph{Filtering:} We select flights between specific airport pairs. Then, we remove duplicate entries,and filter out erroneous data points.
    \item \emph{Normalization \& Resampling:} The surviving trajectories are resampled at uniform time intervals, and key features such as latitude, longitude, altitude, and time since departure are extracted and normalized.
    \item \emph{Clustering:} We group similar flight patterns to support class-conditional generation and targeted trajectory synthesis.
\end{itemize}

\textbf{EUROCONTROL R\&D Archive.} This database serves as a supplementary data source, providing detailed metadata on commercial flights within the EUROCONTROL Network Manager area. While it offers useful flight logs and airspace traversal information, its utility is limited by two factors: restricted availability (four months per year with a two-year delay) and significantly lower sampling rates. These limitations make it less suitable for high-resolution trajectory generation, particularly for critical phases such as takeoff and landing. Nevertheless, the EUROCONTROL data proves valuable for cross-referencing flight information and validating trajectory patterns during the preprocessing of OpenSky data.

\subsection{Model Architecture}

\begin{figure}[!ht]
    \centering
    \includegraphics[width=\columnwidth]{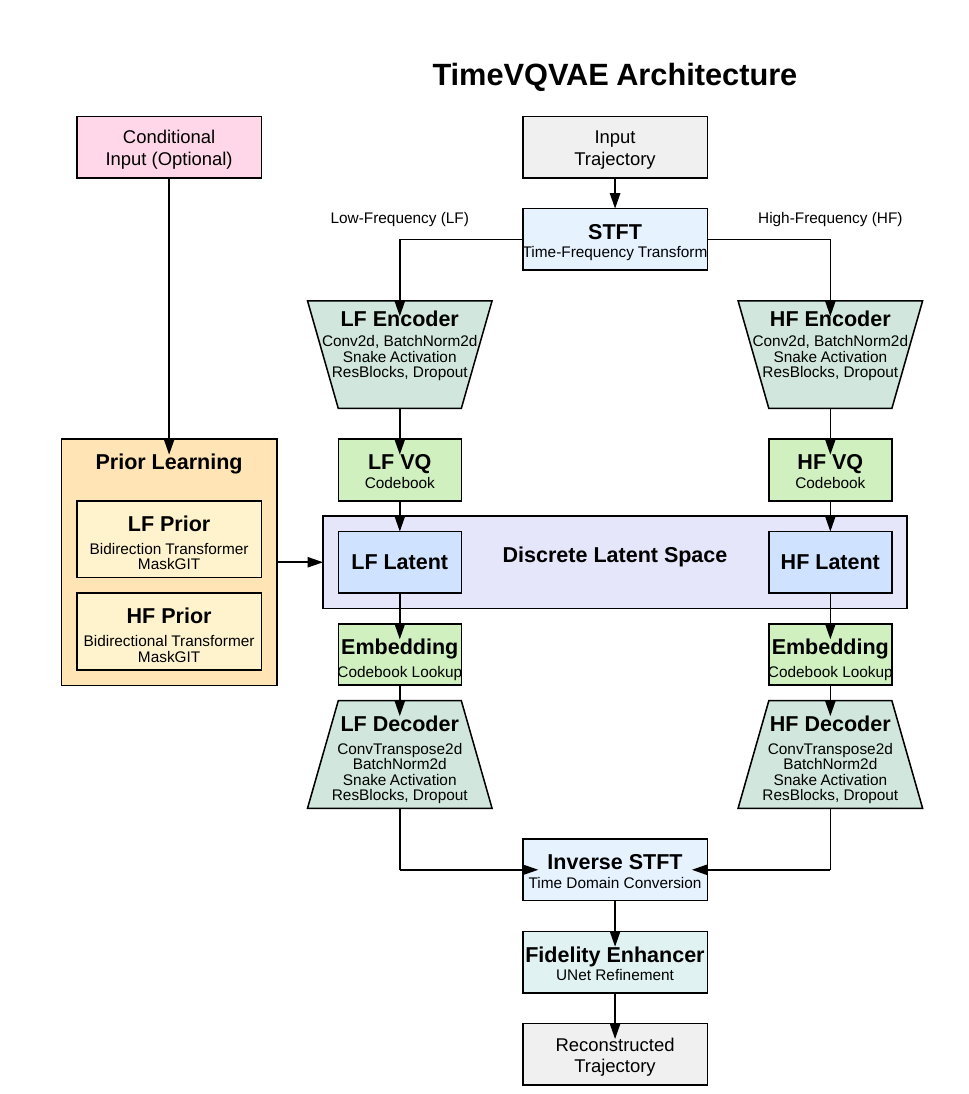}
    \caption{TimeVQVAE Architecture for Generating Synthetic Aircraft Trajectories.}
    \label{fig:architecture}
\end{figure}

Our approach uses a TimeVQVAE architecture \cite{lee_vector_2023} to learn compressed, discrete representations of flight trajectories and to generate new samples.  It integrates advanced machine learning techniques to capture and reproduce complex temporal patterns in flight data. Below, we describe its main components, as illustrated in figure~\ref{fig:architecture}.

\subsubsection{Time-Frequency Transformation}

We apply the Short-Time Fourier Transform (STFT) to convert input trajectories into a time-frequency representation. This transformation helps the model to capture both global trends (low-frequency components) and local, fine-grained patterns (high-frequency components) in the flight data.

\subsubsection{Dual Encoding Pathway}
After the STFT, we split each transformed trajectory into low-frequency (LF) and high-frequency (HF) bands. Each band is processed by a separate encoder, \(E_{LF}\) and \(E_{HF}\), which consist of convolutional layers, residual blocks, and downsampling:
\begin{align}
    z_{LF} &= E_{LF}(\mathcal{P}_{LF}(\mathrm{STFT}(\tau))),\notag\\
    z_{HF} &= E_{HF}(\mathcal{P}_{HF}(\mathrm{STFT}(\tau))),
\end{align}
where \(\mathcal{P}_{LF}\) and \(\mathcal{P}_{HF}\) perform zero-padding on the respective frequency portions, and \(\tau\) is the original time series. This bifurcation allows the model to learn separate latent features for global trajectory shapes and localized movements.

\subsubsection{Vector Quantization}
Within each encoder path, we use the Vector Quantized Variational Autoencoder (VQ-VAE) \cite{van_den_oord_neural_2017}  technique to  create discrete latent representations. For the quantization process, we uses nearest neighbor search in the embedding space. Specifically, for each latent vector \(z^i\):
\[
z_q^i = e_k,\quad k = \arg \min_j \left\lVert z^i - e_j\right\rVert_2,
\]
where $e_j$ is the embedding vector in the codebook, $k$ is the index of the nearest neighbor, and $z_q^i$ is the resulting quantized representation of the trajectory segment.

\subsubsection{Transformer Priors}
Two bidirectional transformer \cite{devlin2018bert} models learn the prior distributions over the LF and HF discrete codes. These transformer-based priors allow the model to capture long-range dependencies and global structures in the trajectory data, which is essential for generating coherent spatial-temporal sequences.

\subsubsection{Dual Decoding Pathway}

Separate decoders $D_{LF}$ and $D_{HF}$ reconstruct the LF and HF components of the trajectory from the quantized representations, using transposed convolutions and residual blocks:
\begin{equation}
\hat{\tau}_{LF} = D_{LF}(z_q^{LF})
\end{equation}
\begin{equation}
\hat{\tau}_{HF} = D_{HF}(z_q^{HF})
\end{equation}

\subsubsection{Dual Decoding Pathway}
Separate decoders, \(D_{LF}\) and \(D_{HF}\), reconstruct the LF and HF components of the trajectory from their quantized representations. Using transposed convolutions and residual blocks, the decoders perform the following transformations:
\begin{equation}
\hat{\tau}_{LF} = D_{LF}(z_q^{LF})
\end{equation}
\begin{equation}
\hat{\tau}_{HF} = D_{HF}(z_q^{HF})
\end{equation}
These reconstructed components can then be combined to recover the complete trajectory in the time domain using Inverse STFT.

\subsection{Training Procedure}

The training procedure for our TimeVQVAE model for aircraft trajectory generation follows a three-stage approach, similar to the original TimeVQVAE implementation, but adapted for the specific characteristics of flight data.

\subsubsection{Stage 1: VQ-VAE Training}

In the initial stage, we train the encoder, decoder, and vector quantizer. The objective is to effectively compress the input trajectories into discrete tokens while minimizing information loss. The training process aims to minimize the total loss function:
\begin{equation}
    \mathcal{L}_{total} = \mathcal{L}_{traj} + \lambda \mathcal{L}_{codebook}
\end{equation}
where $\mathcal{L}_{traj}$ is the reconstruction loss for the trajectory in both time and time-frequency domains \cite{defossezhigh}, $\mathcal{L}_{codebook}$ is the codebook learning loss, that includes a commitment loss (to encourage the encoder to commit to codebook entries) and an exponential moving average loss (to update codebook entries based on encoder outputs). The $\lambda$ parameter is a weighting factor that balances the reconstruction and codebook learning objectives. The codebook is updated using the straight-through estimator \cite{van_den_oord_neural_2017}, ensuring that the gradients are passed through the quantization step during backpropagation. This approach allows for efficient training of the discrete latent representations while maintaining the benefits of vector quantization. We use the Adam optimizer to minimize the total loss function, updating the parameters of the encoder, decoder, and codebook.

\subsubsection{Stage 2: Prior Learning}

Next, we freeze the VQ-VAE components and train the transformer-based prior models to learn the distribution of the discrete tokens.  We adopt a masking approach, where random tokens are masked and then predicted by the transformer. The training loss is the negative log-likelihood of predicting the masked tokens:
\begin{equation}
\mathcal{L}_{\text{mask}} = -\mathbb{E}_s \bigl[\log p_\theta(s^{LF} | s_M^{LF}) + \log p_\phi(s^{HF} \mid s^{LF}, s_M^{HF})\bigr]
\end{equation}
where $\theta$ and $\phi$ are the parameters of the LF and HF transformer models, respectively. We implement class-conditional training by appending class tokens to the input sequences during training, with a probability \(p_{\text{uncond}}\) of using a null class token for unconditional training. The model employs MaskGIT \cite{chang_maskgit_2022} with an iterative decoding process for efficient and coherent generation of latent codes. This approach allows for parallel generation of latent codes, leading to faster sampling compared to autoregressive methods, while maintaining global coherence in the generated sequences.

\subsubsection{Stage 3: Fidelity Enhancement}

Finally, we train an optional fidelity enhancer to refine the generated trajectories. After decoding with the VQ-VAE, we pass the resulting samples through this module to improve coherence and realism. The loss function combines a reconstruction term between enhanced and real samples with a perceptual loss computed using features from a pre-trained Fully Convolutional Network (FCN). This approach thus aligns the generated trajectories more closely with observed flight patterns.

\subsection{Trajectory Generation}

Generation employs a double-pass iterative decoding process over T iterations. Starting with fully masked sequences, the LF transformer first predicts probabilities for all masked LF tokens, selecting and unmasking those with highest confidence. Once the LF codes stabilize, the HF transformer uses this sequence to similarly predict and unmask HF tokens. Both decoders then reconstruct their respective components, which are merged in the time-frequency domain. This split approach helps maintain consistency across frequency scales, balancing broad flight patterns with finer-grained details in the generated trajectories.

\section{Evaluating Synthetic Trajectories} \label{sec-evaluation}

The true effectiveness of a generative model cannot be captured by a single metric alone. In this section, we outline a set of complementary approaches for evaluating how well generated trajectories align with real-world flight data. We begin by describing two widely used quality measures followed by statistical metrics aimed at capturing detailed distributional and temporal nuances. We also integrate a domain-specific “flyability” assessment using an open-source air traffic simulator to check whether these trajectories remain operationally feasible. Finally, we employ various visualization techniques to gain further insights into the strengths and limitations of our approach.

\subsection{Quality Metrics}

We employ two well-known metrics—Fréchet Inception Distance (FID) and Inception Score (IS)—to gauge various aspects of our generated trajectories, from how they match the distribution of real data to their overall diversity.

\subsubsection{Fréchet Inception Distance (FID)}
FID quantifies the difference in feature distributions between real and synthetic data. We adapt this metric to aircraft trajectories by extracting features via a pre-trained Fully Convolutional Network (FCN) or the ROCKET method \cite{dempster_rocket_2020}. FID is then computed by comparing the means (\(\mu_r\), \(\mu_g\)) and covariances (\(\Sigma_r\), \(\Sigma_g\)) of these features:
\begin{equation}
\text{FID} = \|\mu_r - \mu_g\|^2 + \text{Tr}\bigl(\Sigma_r + \Sigma_g - 2\sqrt{\Sigma_r\,\Sigma_g}\bigr)
\end{equation}
where $\mu_r$ and $\Sigma_r$ are the mean and covariance of the real data features, and $\mu_g$ and $\Sigma_g$ are those of the synthetic data. The first term of the equation quantifies the difference in the average features between the real and generated data, while the second term captures the variability and relationships between different features.

Lower FID values suggest that the generated trajectories capture some of the principal statistical characteristics and temporal patterns found in the real dataset. However, FID may not capture rare events or specific patterns within the data, and its effectiveness depends on the representativeness of the feature extraction model. Therefore, FID should be employed in conjunction with other metrics for a comprehensive evaluation of synthetic data quality.

\subsubsection{Inception Score (IS)}
The Inception Score aims to characterize both the \emph{realism} and \emph{diversity} of generated samples. It computes the Kullback-Leibler (KL) divergence between the conditional class distribution \(p(y\mid x)\) and the marginal class distribution \(p(y)\) across all generated samples:
\begin{equation}
\text{IS} = \exp\bigl(\mathbb{E}_x \bigl[\mathrm{KL}(p(y \mid x)\| p(y))\bigr]\bigr).
\end{equation}
Here, $x$ is a generated trajectory, $y$ denotes the class label, and $p(y|x)$ and $p(y)$ are the conditional and marginal class distributions, respectively. We obtain $p(y|x)$ by passing the generated trajectory through a pre-trained Fully Convolutional Network and calculating the class probabilities.

Higher IS values suggest greater alignment with real data variety and more pronounced class distinctions. However, IS alone may underrepresent subtler or less frequent flight patterns. Its interpretation can also be less intuitive compared to distance-based metrics like FID. Therefore, as with FID, we treat IS as one part of a broader evaluation.

\subsection{Statistical Metrics}

To capture how closely the generated trajectories replicate not only the average behavior but also its distributional and temporal characteristics, we use the following statistical metrics:

\subsubsection{Marginal Distribution Difference (MDD)}
MDD quantifies the similarity between the marginal distributions of real and synthetic datasets \cite{ni_sig-wasserstein_2021}. It is defined as the average absolute difference between kernel density estimates (KDEs) for real and generated samples:
\begin{equation}
\text{MDD} = \frac{1}{n} \sum_{i=1}^{n} \bigl| f_{\mathrm{real}}(x_i) - f_{\mathrm{gen}}(x_i) \bigr|.
\end{equation}
Smaller MDD values indicate closer alignment in overall characteristics, such as altitude profiles and geographical coordinates, making it crucial for reflecting typical air traffic patterns.

\subsubsection{Autocorrelation Difference (ACD)}
ACD measures how well the synthetic data captures the temporal dependencies present in real data \cite{lai_modeling_2018}. We compute ACD using the mean absolute difference between the autocorrelation coefficients of real and synthetic trajectories:
\begin{equation}
\text{ACD} = \frac{1}{n} \sum_{i=1}^{n} \bigl|\rho_{\mathrm{real}}(k) - \rho_{\mathrm{gen}}(k)\bigr|,
\end{equation}
where \(\rho(k)\) is the autocorrelation coefficient at lag \(k\), computed using the Pearson correlation formula\cite{freedman2007statistics}. A lower ACD suggests more accurate modeling of temporal patterns and rhythms found in actual flight data. This includes representing various flight phases such as climb, cruise, and descent, as well as capturing any cyclical patterns that might exist in air traffic data.

\subsubsection{Skewness Difference (SD)}
Skewness Difference helps compare asymmetry in distributions of real and synthetic data:
\begin{equation}
\text{SD} = \bigl|\text{Skew}_{\mathrm{real}} - \text{Skew}_{\mathrm{gen}}\bigr|,
\end{equation}
where the skewness values are calculated using the third standardized moment of the distribution.
A low SD indicates that the synthetic data models the skewness of the real data, capturing phenomena such as uneven distributions of flight altitudes or speeds. These asymmetries may arise from factors such as air traffic control practices, aircraft performance characteristics, or preferred flight levels.

\subsubsection{Kurtosis Difference (KD)}
KD Compares the "tailedness" of the distributions between real and synthetic data, providing insight into how well extreme values are represented: 
\begin{equation}
\text{KD} = \bigl|\text{Kurt}_{\mathrm{real}} - \text{Kurt}_{\mathrm{gen}}\bigr|,
\end{equation}
where the kurtosis values are calculated using the fourth standardized moment of the distribution. A low KD indicates that the synthetic data captures the kurtosis of the real data, including the presence of extreme values or outlier events. These could include rare but significant flight patterns, such as emergency diversions, unusual altitude profiles, or extreme weather-related deviations.

\subsection{Domain-Specific Metric: Flyability Assessment}

To examine whether the generated synthetic trajectories can be flown, we use the open-source BlueSky simulator \cite{hoekstra_bluesky_2016}. We feed the synthetic trajectories into the simulator, which converts them into a series of waypoints (latitude, longitude, altitude). The simulator then executes these trajectories, producing the closest "flyable" paths based on its physics model and operational constraints. To quantify the difference, we compute various distance metrics between the original and simulated trajectories, where lower distances indicate closer alignment with physical and operational constraints. This approach provides a spectrum measure of "flyability," offering more insight than a binary pass/fail evaluation.

We utilize a range of trajectory distance metrics to evaluate the similarity between generated and simulated trajectories:
\begin{itemize}
    \item Symmetric Segment-Path Distance (SSPD): Provides a balanced measure of similarity between two trajectories by considering both point-to-point and point-to-path distances \cite{besse_review_2016}.
    \item One-Way Distance (OWD): Uses a grid to efficiently compute distances for trajectories of varying lengths or sampling rates \cite{lin_shapes_2005}.
    \item Hausdorff Distance: Measures the maximum deviation between trajectories, providing a measure of the worst-case dissimilarity.
    \item Fréchet Distance: Accounts for the continuity of trajectories, capturing similarities in the overall shape and progression of flight paths.
    \item Discrete Fréchet Distance: A computationally efficient alternative to the continuous Fréchet distance, providing a good approximation for discretely sampled trajectories.
    \item Dynamic Time Warping (DTW): Allows for non-linear alignment between trajectories, useful for comparing flights with similar spatial paths but different speed profiles.
    \item Edit Distance with Real Penalty (ERP): Combines edit distance and real-valued distances, robust to noise and sampling variations \cite{chen_marriage_2004}.
    \item Edit Distance on Real sequence (EDR): Compares trajectories with less sensitivity to outliers \cite{chen_robust_2005}.
\end{itemize}

\subsection{Visual Inspection}

Beyond numerical metrics, we use a range of visualization techniques—such as PCA, t-SNE, time series plots, correlation heatmaps, distribution plots, and trajectory maps—to examine the realism and coherence of synthetic flight trajectories. These visual methods highlight underlying structures, clusters, and potential outliers in the generated data. By complementing quantitative evaluations with detailed visual assessments, we gain a more comprehensive view of the model’s performance and can more readily identify areas that require refinement.

\section{Results and Discussion}

To evaluate the TimeVQVAE model’s performance, we apply the evaluation methods outlined in Section \ref{sec-evaluation} using a representative route from Amsterdam (EHAM) to Milan (LIMC). The dataset is divided into training and validation sets, with the model trained on the former. The generated trajectories are then compared to those in the validation set to assess accuracy and generalization. This approach mitigates overfitting and assesses generalization.

\subsection{Quality \& Statistical Metrics}

Table~\ref{tab:combined_performance_metrics} summarizes both quality and statistical metrics, comparing TimeVQVAE to a baseline TCVAE. Overall, the TimeVQVAE outperforms TCVAE across all metrics. Key observations include:
\begin{itemize} 
    \item \textbf{FID}: TimeVQVAE attains a notably low FID of 0.0029, versus 0.7944 for TCVAE, suggesting closer alignment with the real data distribution.
    \item \textbf{IS}: TimeVQVAE’s higher IS (3.4049 ± 0.2660 compared to TCVAE’s 2.1514 ± 0.1283) suggests better quality and diversity in the generated samples. However, the relatively low scores on the typical IS scale (1-10) for both models may reflect inherent similarities in flight paths due to standardized routes and procedures. 
    \item \textbf{MDD}: TimeVQVAE’s lower MDD (0.0512 vs. 0.1789) indicates a better approximation of the overall distribution of real trajectory data. 
    \item \textbf{ACD}: While TimeVQVAE demonstrates a significantly lower ACD (5.6334 vs. 289.3323), both models show room for improvement in capturing fine-grained temporal structures. 
    \item \textbf{SD}: TimeVQVAE’s lower SD (0.0554 vs. 0.2253) demonstrates better reproduction of skewness characteristics, which is critical for capturing asymmetries in flight parameters. 
    \item \textbf{KD}: The lower KD for TimeVQVAE (0.1065 vs. 0.3164) indicates a better representation of extreme trajectory characteristics. 
\end{itemize}
\begin{table}[!htbp]
\centering
\scriptsize  
\setlength{\tabcolsep}{6pt} 
\renewcommand{\arraystretch}{1.2} 
\caption{Quality and Statistical Metrics}
\label{tab:combined_performance_metrics}
\begin{tabular}{lcc}
\toprule
\rowcolor{blue!15} 
\textbf{Metric}  & \textbf{TCVAE} & \textbf{TimeVQVAE} \\
\midrule
\rowcolor{gray!5} 
\textbf{FID} ($\downarrow$) & 0.7944 & \textbf{0.0029} \\
\rowcolor{gray!20} 
\textbf{IS} ($\uparrow$) & 2.1514 ± 0.1283 & \textbf{3.4049 ± 0.2660} \\
\rowcolor{gray!5} 
\textbf{MDD} ($\downarrow$ )& 0.1789 & \textbf{0.0512} \\
\rowcolor{gray!20} 
\textbf{ACD} ($\downarrow$) & 289.3323 & \textbf{5.6334} \\
\rowcolor{gray!5} 
\textbf{SD} ($\downarrow$) & 0.2253 & \textbf{0.0554} \\
\rowcolor{gray!20} 
\textbf{KD} ($\downarrow$) & 0.3164 & \textbf{0.1065} \\
\bottomrule
\end{tabular}

\end{table}

\subsection{Flyability Assessment}
\begin{figure}[!ht]
    \centering
    \includegraphics[width=\columnwidth]{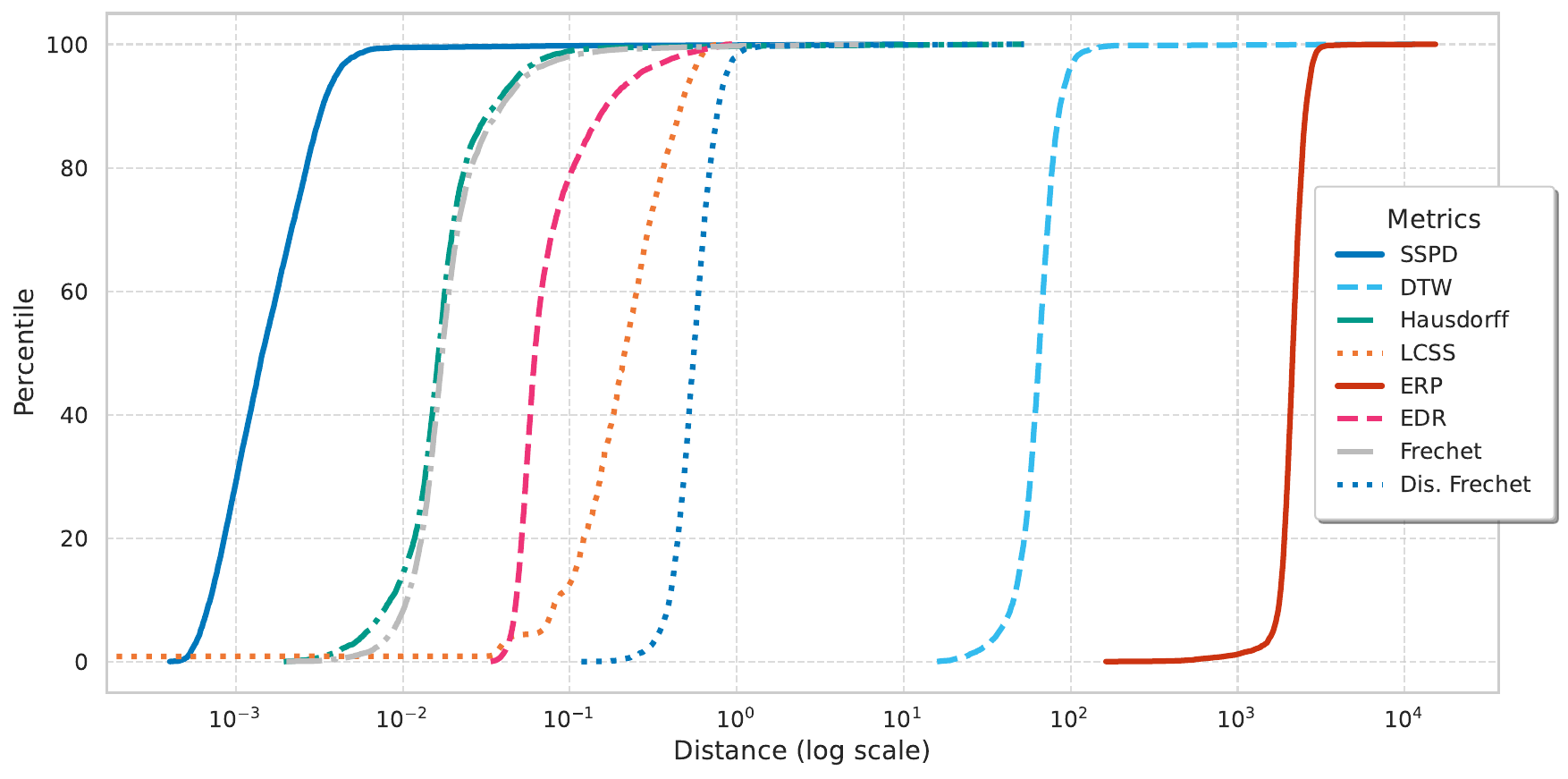}
    \caption{Percentile Plots of distance metrics between synthetic trajectories and their simulated counterparts.}
    \label{fig:flyability}
\end{figure}

Figure \ref{fig:flyability} presents percentile plots for the eight distance metrics, offering insights into the flyability of the TimeVQVAE-generated trajectories. Key observations from the flyability assessment are as follows:
\begin{itemize} 
    \item \textbf{SSPD}: Demonstrates the best performance, with nearly 80\% of trajectories having a distance less than 0.01. This indicates a close overall shape and path matching between the synthetic and simulated trajectories.
    \item \textbf{Hausdorff and Fréchet distances}: Exhibit similar distributions, with approximately 60\% of trajectories having distances less than 0.1. These metrics, which are sensitive to the maximum deviation between trajectories, suggest that while most of the trajectory aligns well, there may be some points of significant deviation.
    \item \textbf{LCSS and EDR}: Show higher distances, with only about 20\% of trajectories having distances less than 0.1. This suggests that while the overall shape may be similar, there are notable differences in the detailed sequence of points between the synthetic and simulated trajectories.
    \item \textbf{DTW and ERP}: Record the highest distances, with almost all trajectories showing distances greater than 1. This indicates significant temporal misalignments or structural differences between the synthetic and simulated trajectories. 
\end{itemize}

Overall, these findings indicate that the TimeVQVAE generates generally flyable trajectories, but there are some discrepancies between the generated and simulated paths, particularly in temporal aspects and fine-grained details.

\begin{figure}[!ht]
    \centering
    \includegraphics[width=\columnwidth]{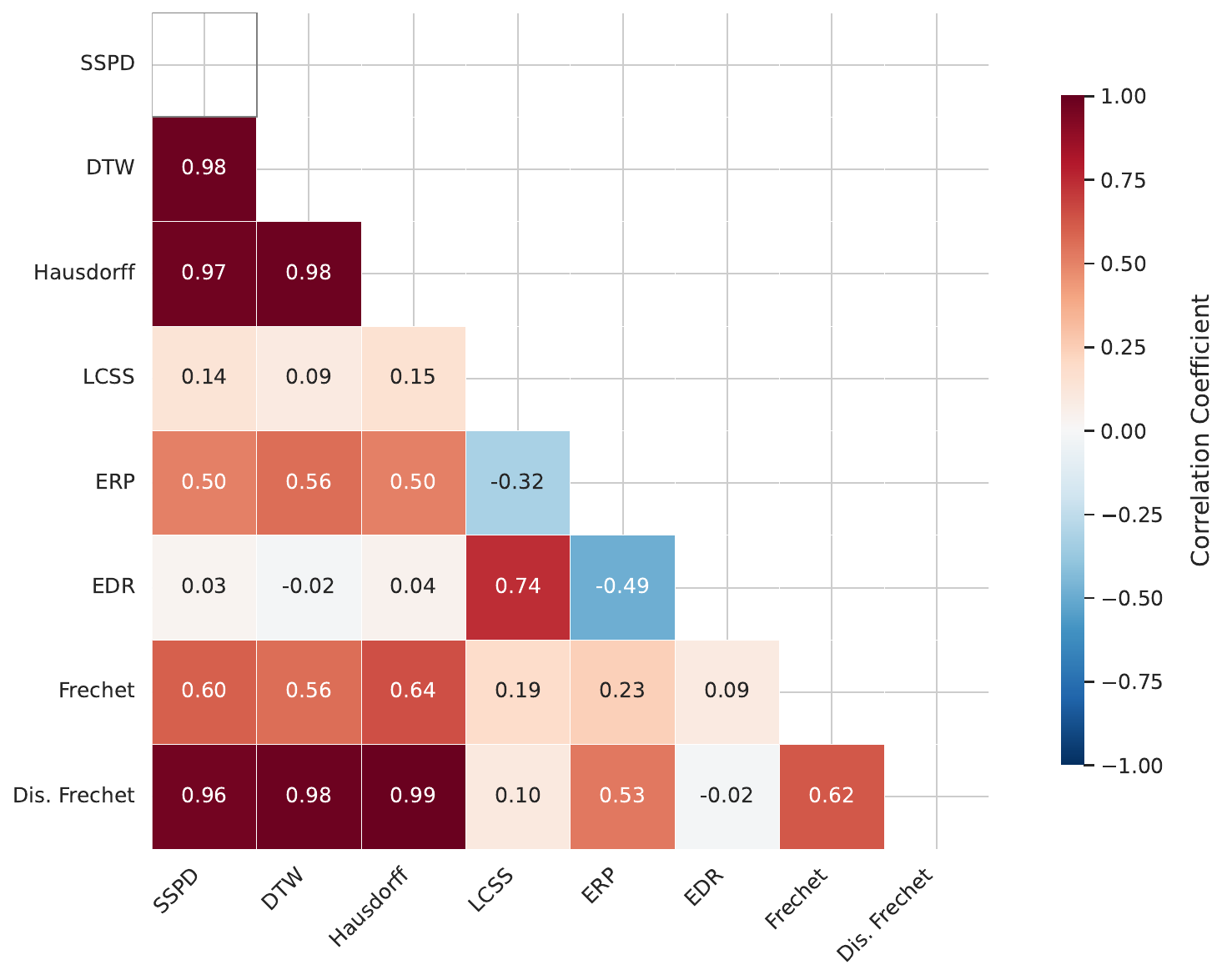}
    \caption{Correlation Heatmap for Euclidean Distance Metrics.}
    \label{fig:correlation_heatmap_euclidean}
\end{figure}

Figure \ref{fig:correlation_heatmap_euclidean} presents a correlation heatmap for the distance metrics, offering insights into their interrelationships. The heatmap reveals the following:
\begin{itemize} 
    \item Strong positive correlations (0.96-0.99) between SSPD, DTW, Hausdorff, and Discrete Fréchet distances, suggesting that these metrics capture similar aspects of trajectory differences. 
    \item Moderate correlations (0.56-0.64) between the Fréchet distance and the above group. 
    \item A strong positive correlation (0.74) between LCSS and EDR, but weak or negative correlations with other metrics, indicating that these metrics may capture unique aspects of trajectory similarity. 
\end{itemize}

\subsection{Visual Inspection}

\subsubsection{Principal Component Analysis (PCA)}
\begin{figure}[!ht]
    \centering
    \includegraphics[width=\columnwidth]{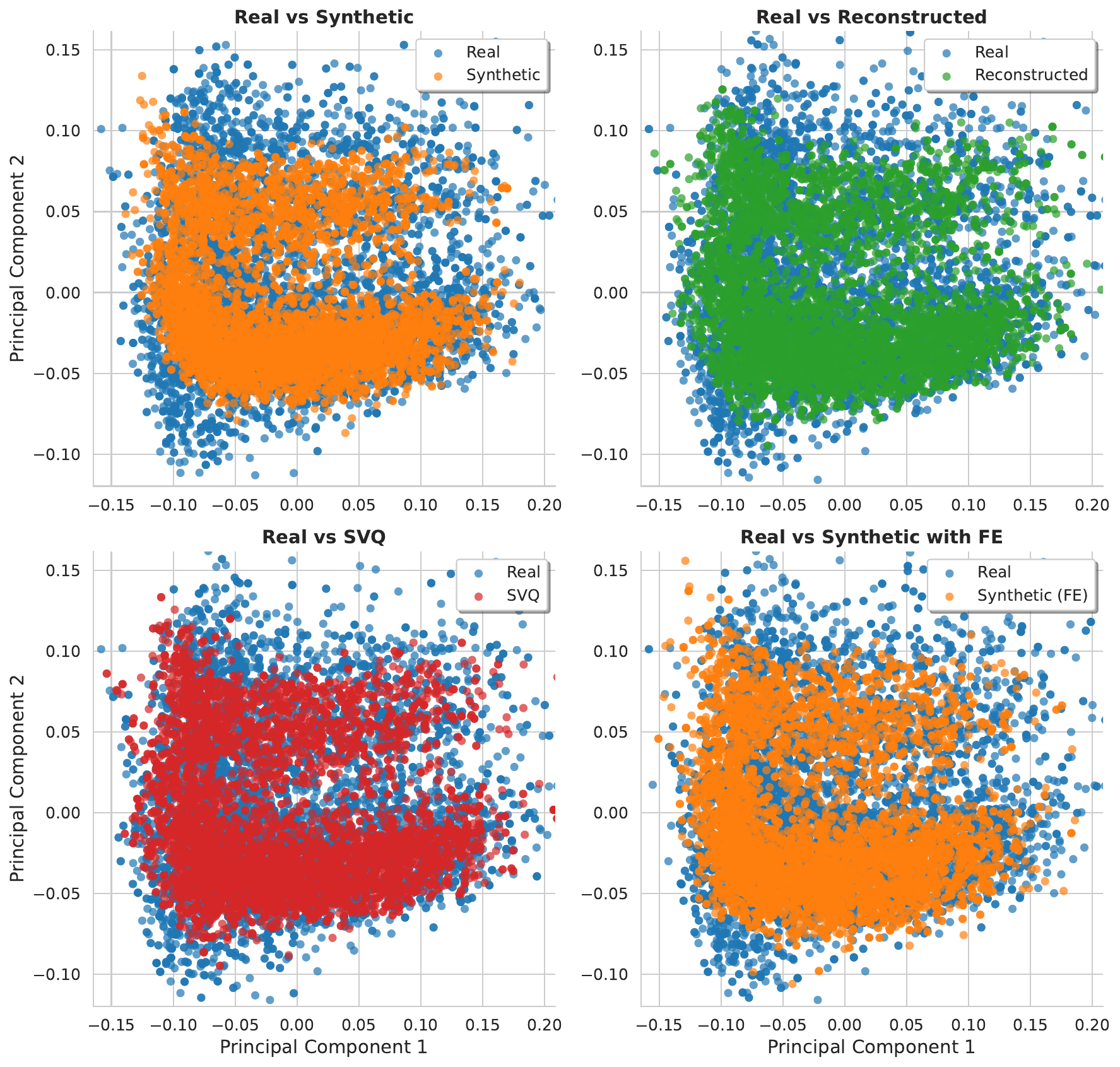}
    \caption{PCA Visualization of Real vs Synthetic Trajectories.}
    \label{fig:pca_visualization}
\end{figure}

We applied PCA to visualize the global structure of real and synthetic trajectories, focusing on the first two principal components, which capture the highest variance. Figure \ref{fig:pca_visualization} shows comparisons between real flight trajectories and synthetic data generated by the TimeVQVAE model, as well as the reconstructed trajectories, latent space representations (through Stochastic Vector Quantization: SVQ), and fidelity-enhanced synthetic data. We observe a substantial overlap between the real and synthetic data, indicating that TimeVQVAE replicates the statistical properties of real flight trajectories. The similar dispersion patterns suggest that the model captures both the central tendencies and variability in flight paths.

\subsubsection{t-Distributed Stochastic Neighbor Embedding (t-SNE)}

\begin{figure}[!ht]
    \centering
    \includegraphics[width=\columnwidth]{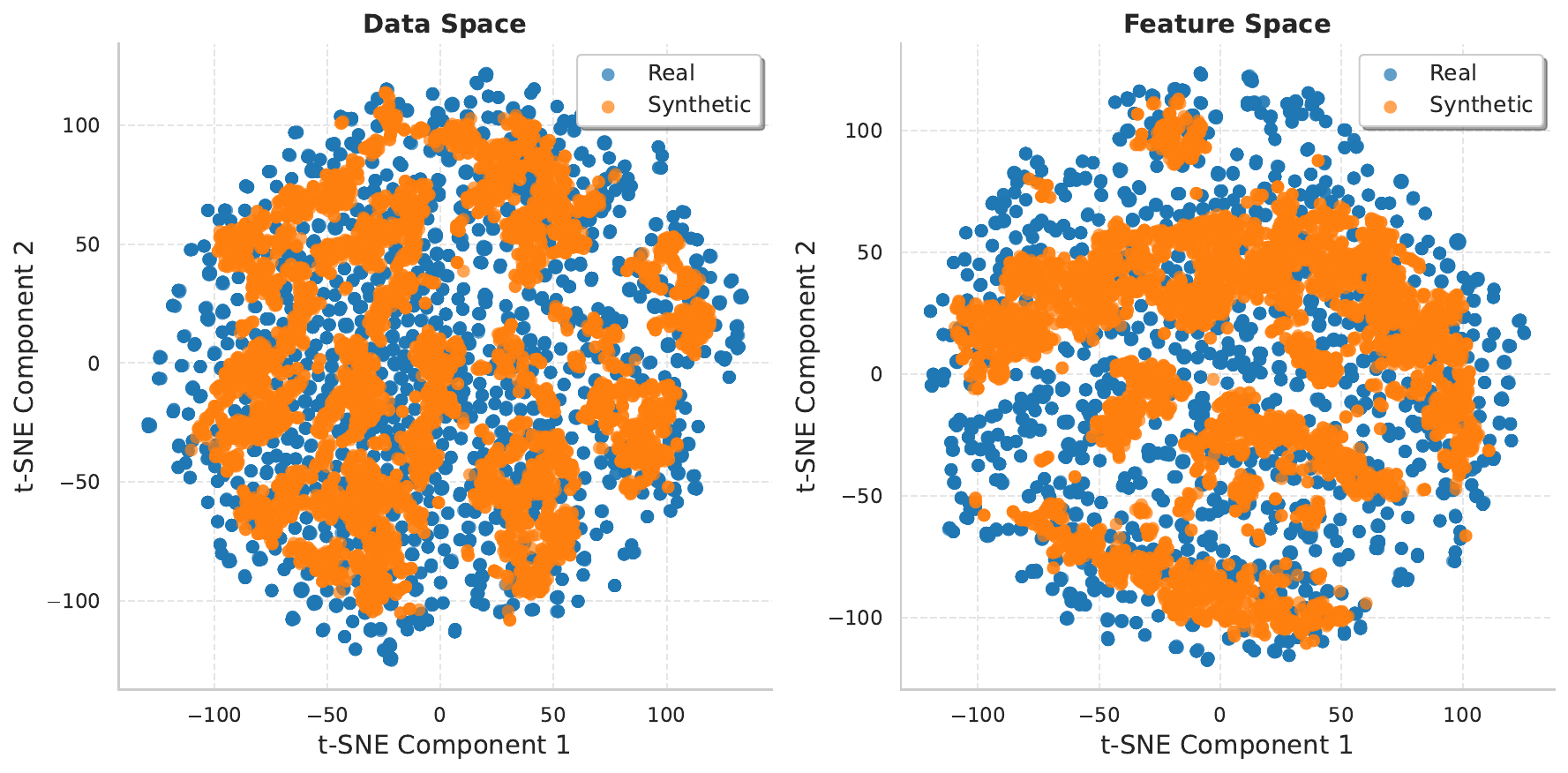}
    \caption{tSNE Visualization}
    \caption{t-SNE Visualization of Real and Synthetic Flight Trajectories in Data Space and Feature Space.}
    \label{fig:tsne_visualization}
\end{figure}

We applied t-SNE to project real and synthetic trajectories into two dimensions, highlighting local structures and differences. Unlike PCA, which offers a global view, t-SNE excels at revealing finer, localized differences, making it particularly useful for assessing the local coherence and realism of synthetic aircraft trajectories. Figure \ref{fig:tsne_visualization} shows t-SNE visualizations for both data space (raw trajectory data) and feature space (ROCKET features).
We observe a significant overlap between real and synthetic points, indicating that TimeVQVAE replicates both the overall characteristics and underlying feature representations of real flight paths. The mixed clusters suggest that the model captures the diversity of flight patterns present in the original dataset. Minor differences in density between real and synthetic points may indicate areas for model refinement.

\subsubsection{Time Series Plots}

\begin{figure}[!ht]
    \centering
    \includegraphics[width=\columnwidth]{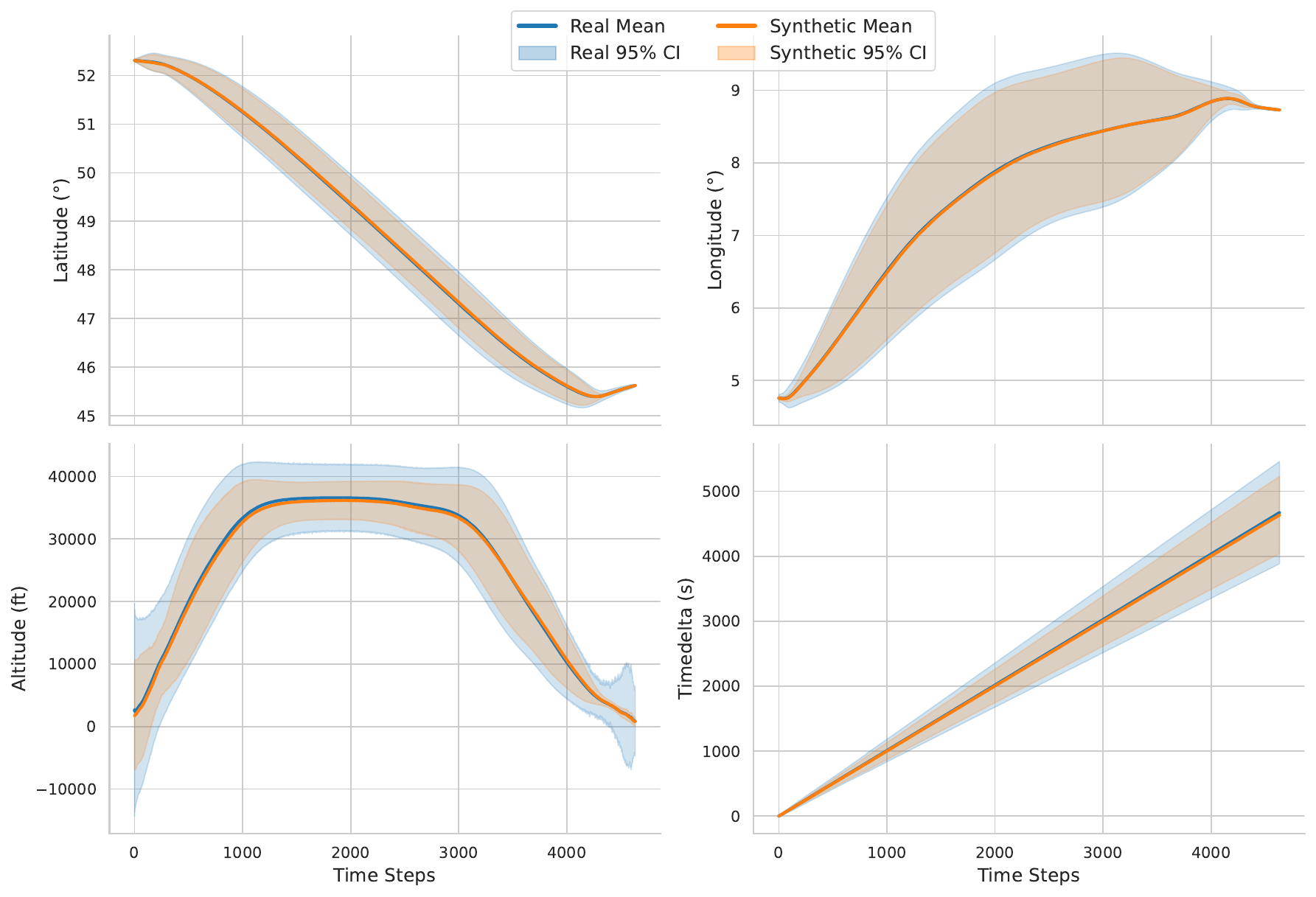}
    \caption{Time Series Plots for Real and Synthetic Trajectories.}
    \label{fig:time_series}
\end{figure}

To evaluate the temporal coherence of synthetic flight trajectories, we analyze time series plots for key flight features: latitude, longitude, altitude, and timedelta. Figure \ref{fig:time_series} illustrates the mean and 95\% confidence interval for each feature over time, comparing real and synthetic trajectories. The synthetic trajectories mirror the overall patterns and temporal dynamics of the real data across all features. Notably, the confidence intervals for the synthetic data are slightly narrower, particularly in the altitude profiles, suggesting that the model generates trajectories with slightly less variability compared to the original dataset. Nevertheless, the synthetic data captures the major trends and temporal evolution of the flight parameters. This analysis demonstrates that TimeVQVAE effectively replicates the temporal characteristics and dynamics of real flight trajectories, while maintaining a balance between fidelity and variability in the generated data.

\subsubsection{Correlation Heatmaps}
\begin{figure}[!ht]
    \centering
    \includegraphics[width=\columnwidth]{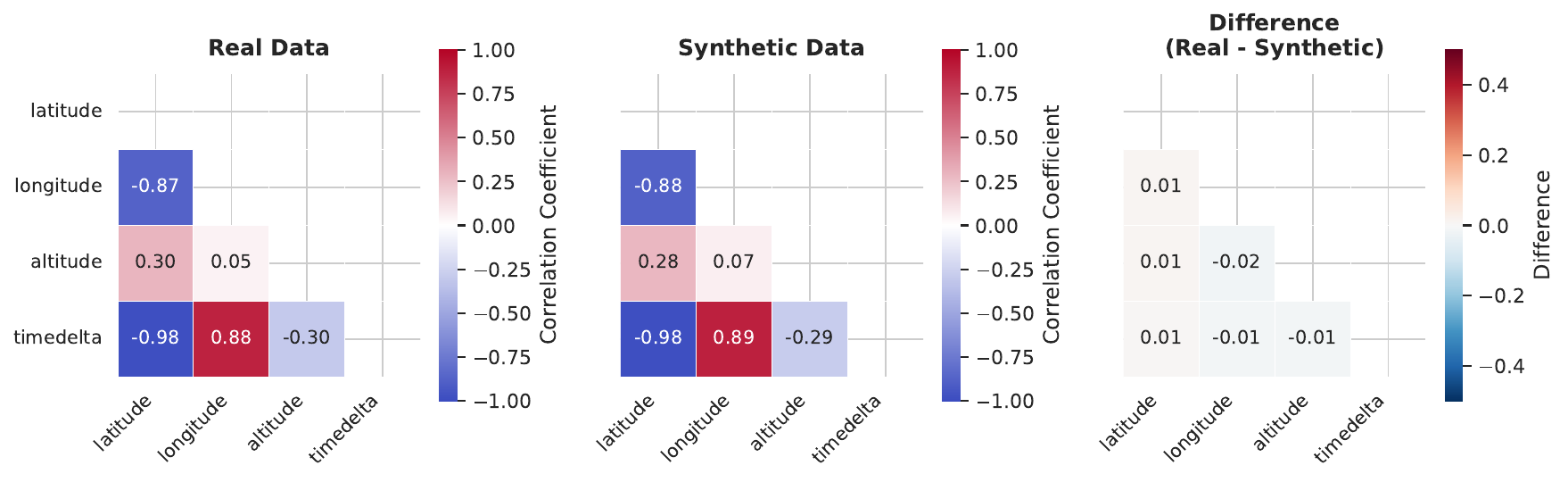}
    \caption{Correlation Heatmaps for Real and Synthetic Trajectories.}
    \label{fig:correlation_heatmaps}
\end{figure}

To assess how well the TimeVQVAE model preserves relationships between flight features, we generate correlation heatmaps for both real and synthetic data. Figure \ref{fig:correlation_heatmaps} presents these heatmaps along with a difference map highlighting discrepancies. The synthetic data captures the overall correlation structure of the real data, reproducing strong correlations like the negative correlation between latitude and longitude (-0.87 vs. -0.88) and the strong positive correlation between latitude and timedelta (-0.98 for both). Weaker correlations, such as those involving altitude, are also well-preserved. The difference heatmap shows minimal discrepancies, with absolute differences of 0.02 or less for all feature pairs, indicating the model's fidelity in maintaining the real data's correlation structure.

\subsubsection{Distribution Plots}
\begin{figure}[!ht]
    \centering
    \includegraphics[width=\columnwidth]{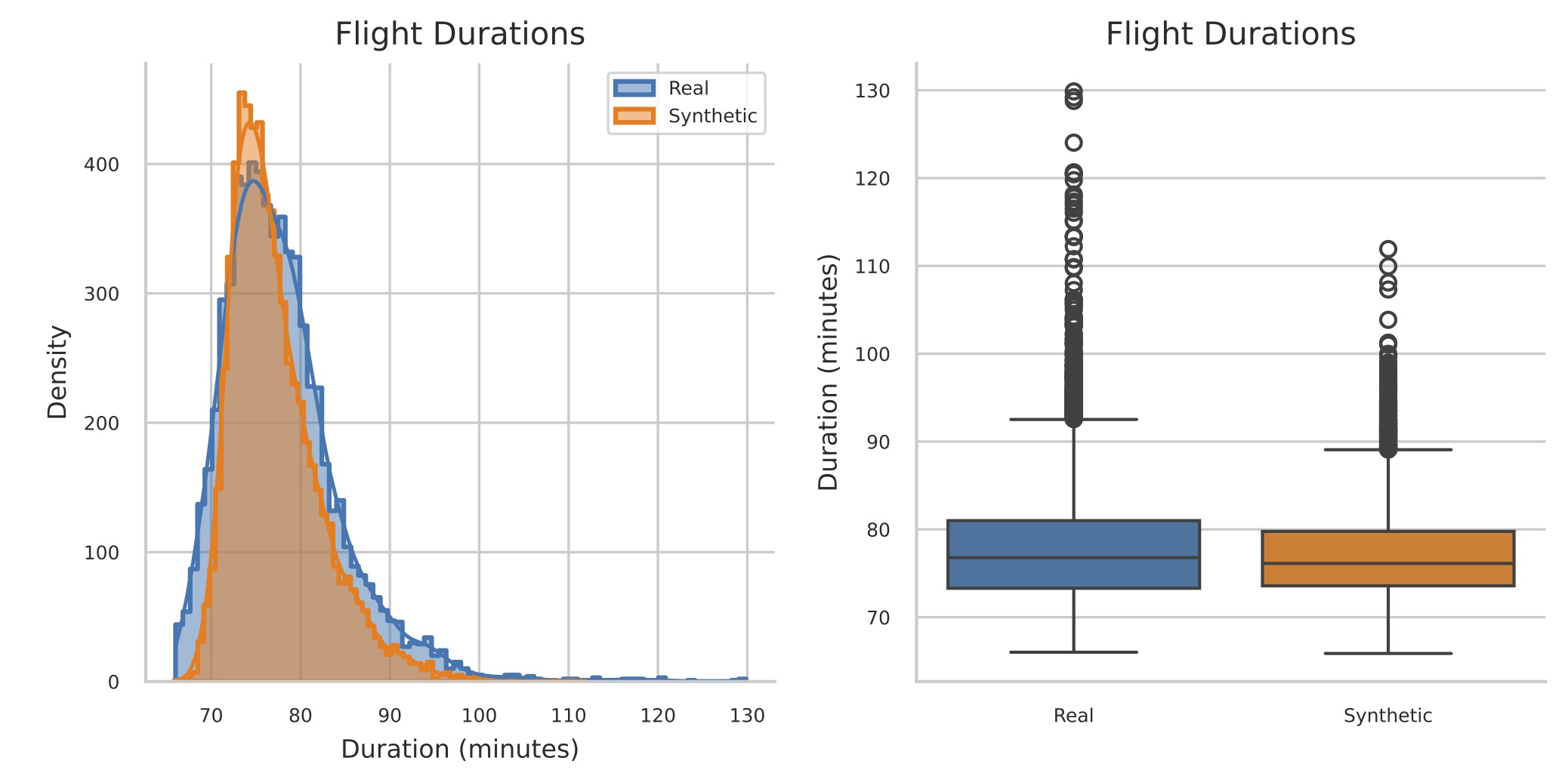}
    \caption{Distribution Plots of Derived Feature: Flight Durations.}
    \label{fig:distribution_plots}
\end{figure}

To evaluate how well the TimeVQVAE model captures the statistical characteristics of real flight data, we examined the distributions a \textit{derived} feature: flight durations. Figure \ref{fig:distribution_plots} displays histograms with kernel density estimates, alongside box plots for each feature, comparing the real data to synthetic data generated by the model. The model replicates the overall shape and central tendency of flight durations, though with slightly reduced variability. Specifically, the synthetic data shows a narrower range of flight durations compared to the real data. 

\subsubsection{Trajectory Visualizations}

\begin{figure}[!ht]
    \centering
    \begin{subfigure}{0.48\columnwidth} 
        \includegraphics[width=\columnwidth]{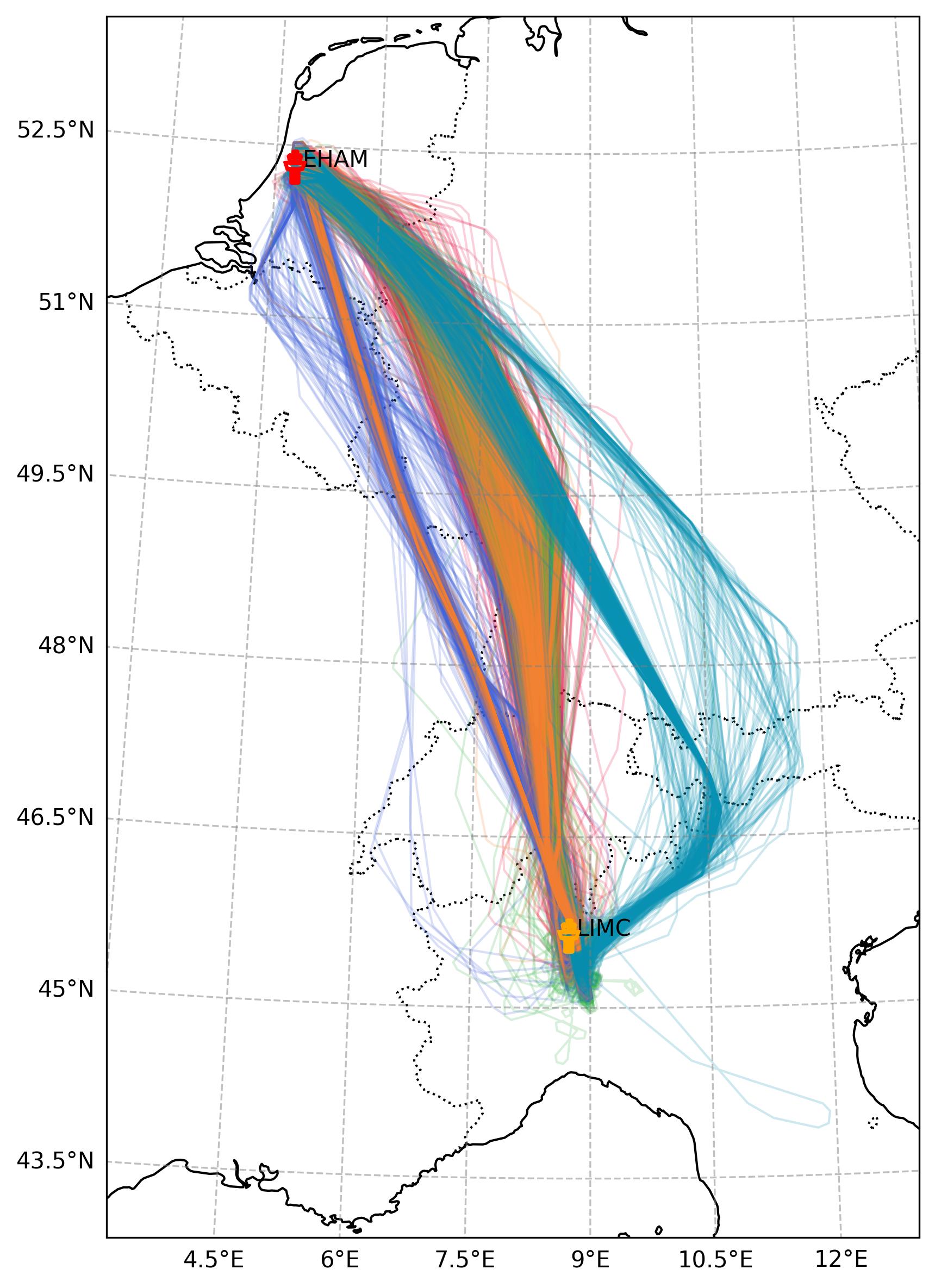}
        \caption{Real Trajectories}
    \end{subfigure}
    \hfill
    \begin{subfigure}{0.48\columnwidth}
        \includegraphics[width=\columnwidth]{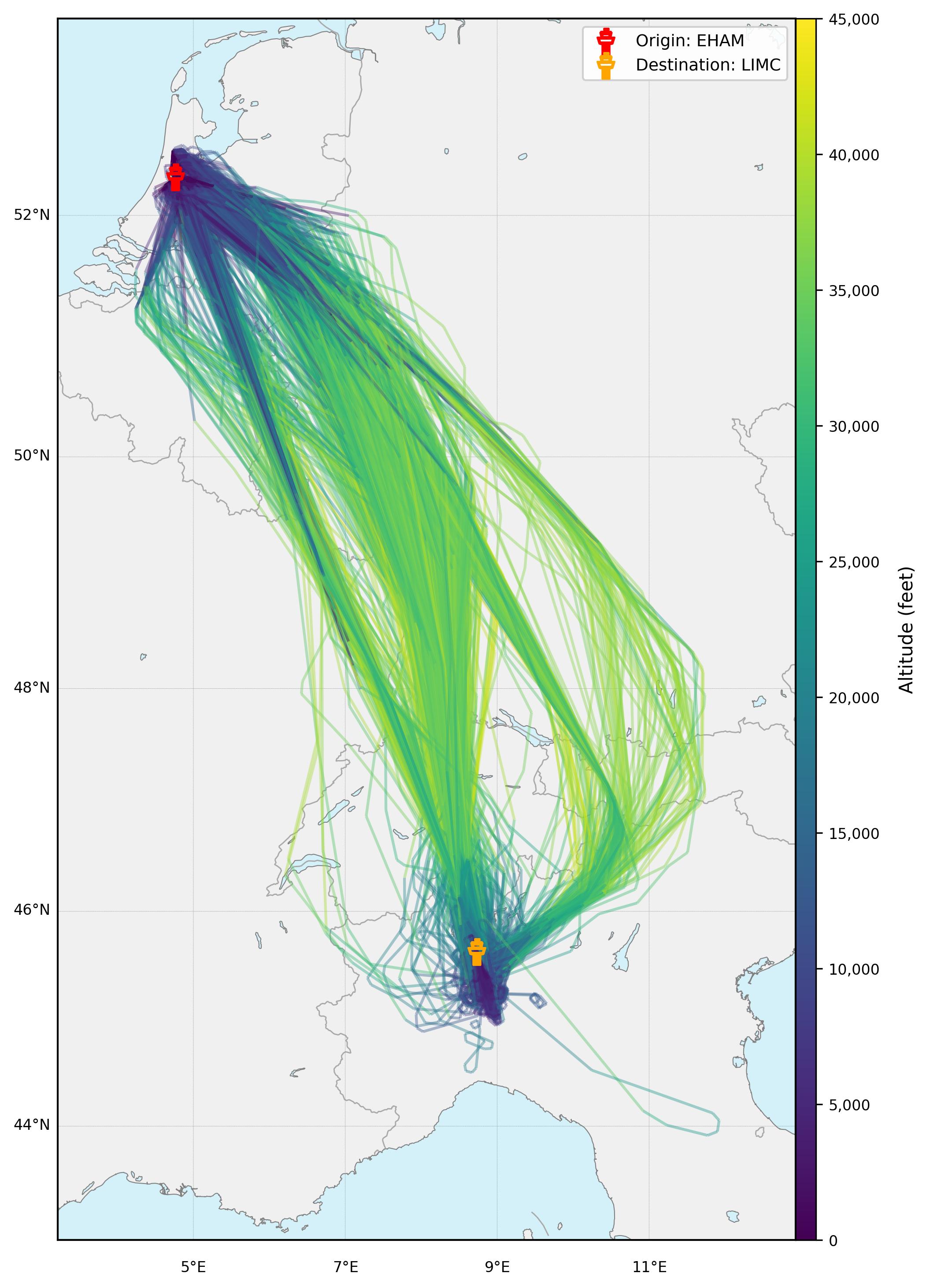}
        \caption{Real Altitude Profile}
    \end{subfigure}
    \vspace{0.2cm}

    \begin{subfigure}{0.48\columnwidth} 
        \includegraphics[width=\columnwidth]{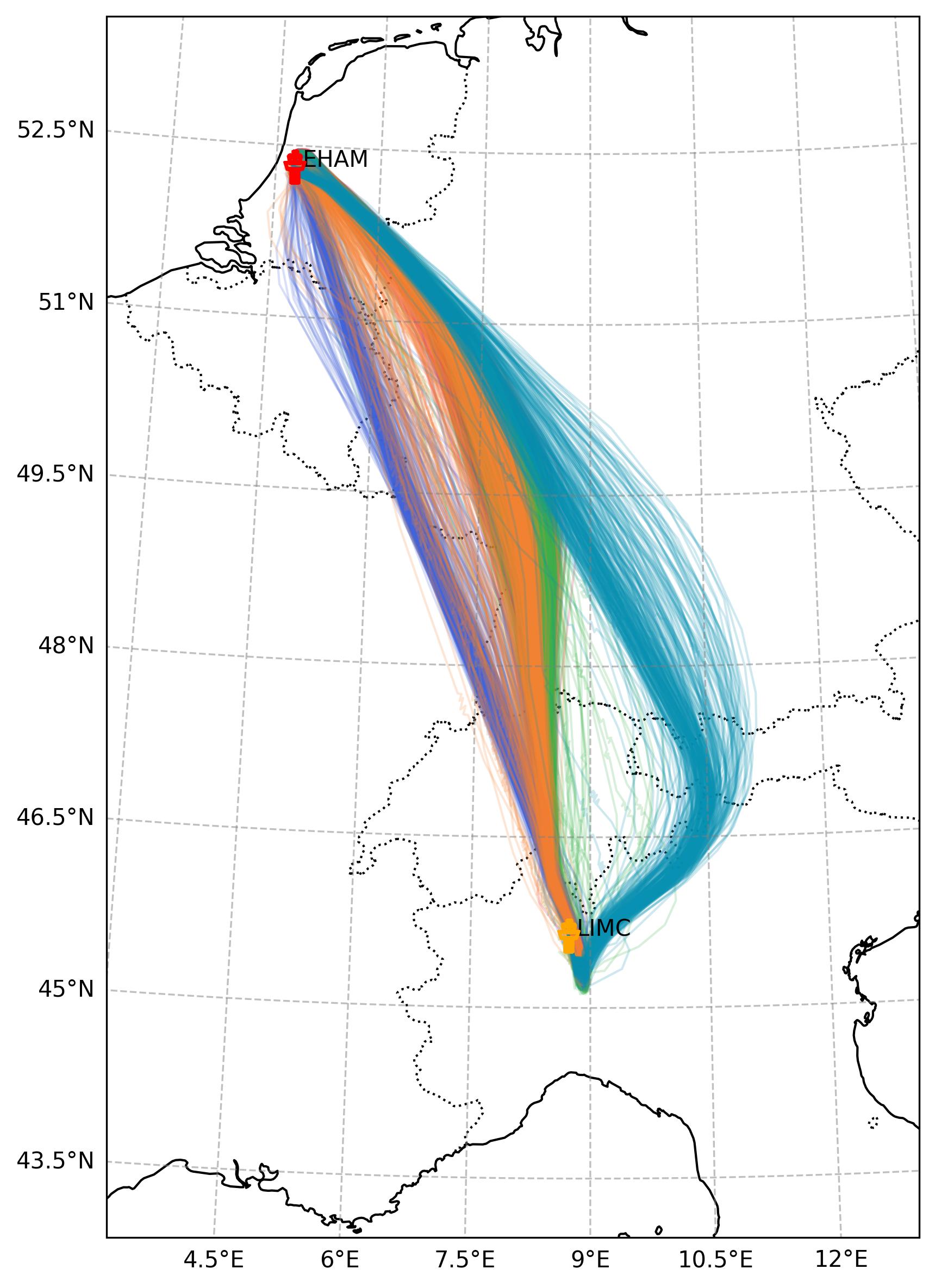}
        \caption{Synthetic Trajectories}
    \end{subfigure}
    \hfill
    \begin{subfigure}{0.48\columnwidth}
        \includegraphics[width=\columnwidth]{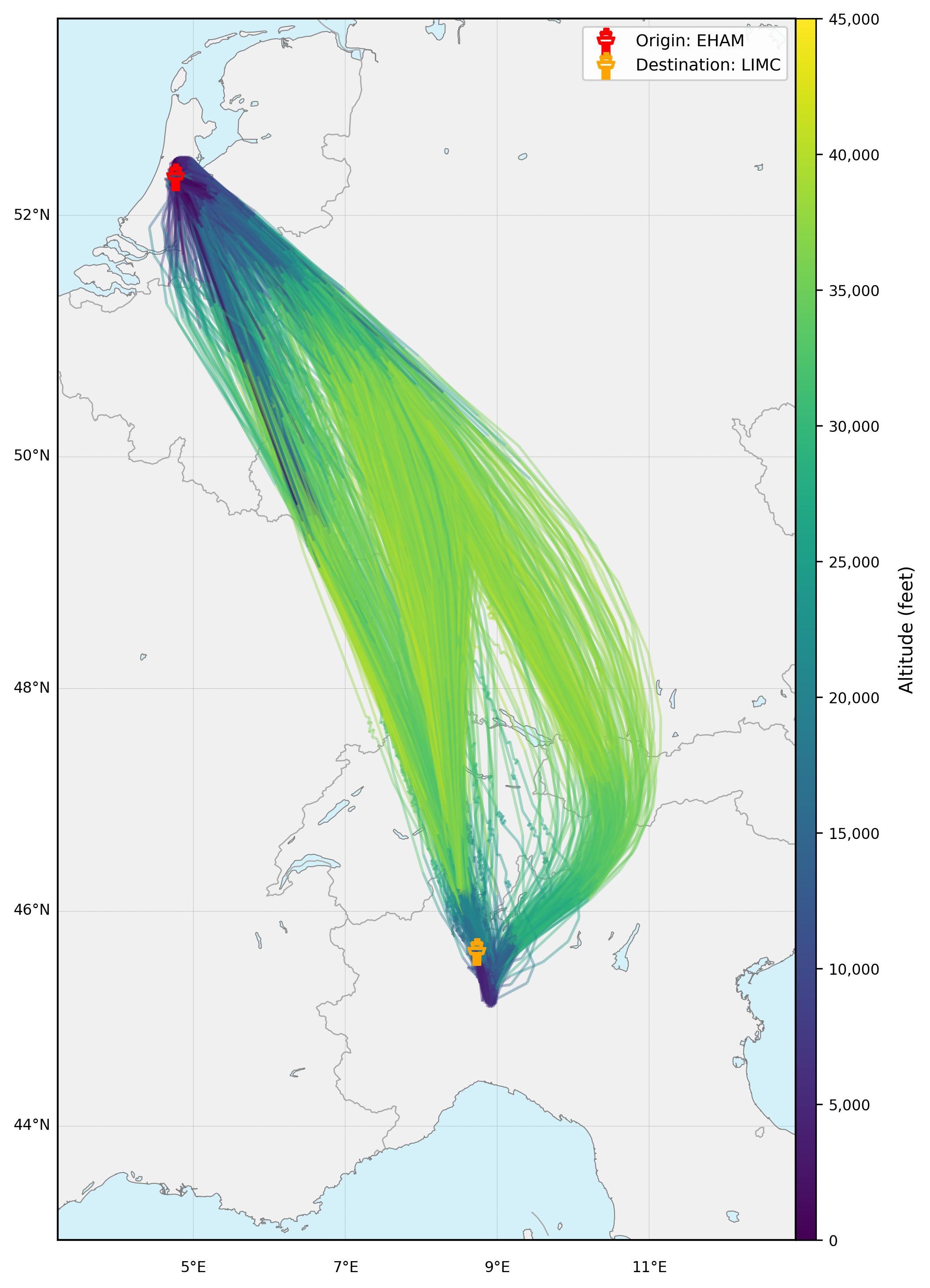}
        \caption{Synthetic Altitude Profile}
    \end{subfigure}
    \caption{Comparison of Real and Synthetic Trajectories (EHAM to LIMC) in Terms of Spatial Routes and Altitude Profiles.}
    \label{fig:trajectory_comparison}
\end{figure}

Figures~\ref{fig:trajectory_comparison} compare real (top row) and synthetic (bottom row) flight trajectories from EHAM to LIMC, with color-coding indicating their respective clusters (i.e., classes). This class-conditional coloring shows how the model can generate diverse, targeted flight paths by focusing on particular trajectory types.

While the synthetic trajectories successfully capture primary route corridors and the broad altitude spectrum—spanning takeoff, cruise, and landing—they exhibit noticeably less variation during the cruise phase. This smoothing effect is in line with the time series observations in Figure~\ref{fig:time_series}, suggesting that the model may underrepresent rare events or outlier behaviors (e.g., turbulence-related altitude adjustments). Overall, the synthetic data reproduces key flight patterns but provides a narrower range of operational dynamics than is observed in real-world trajectories.

\begin{figure}[!ht]
    \centering
    \begin{subfigure}{0.48\columnwidth} 
        \includegraphics[width=\columnwidth]{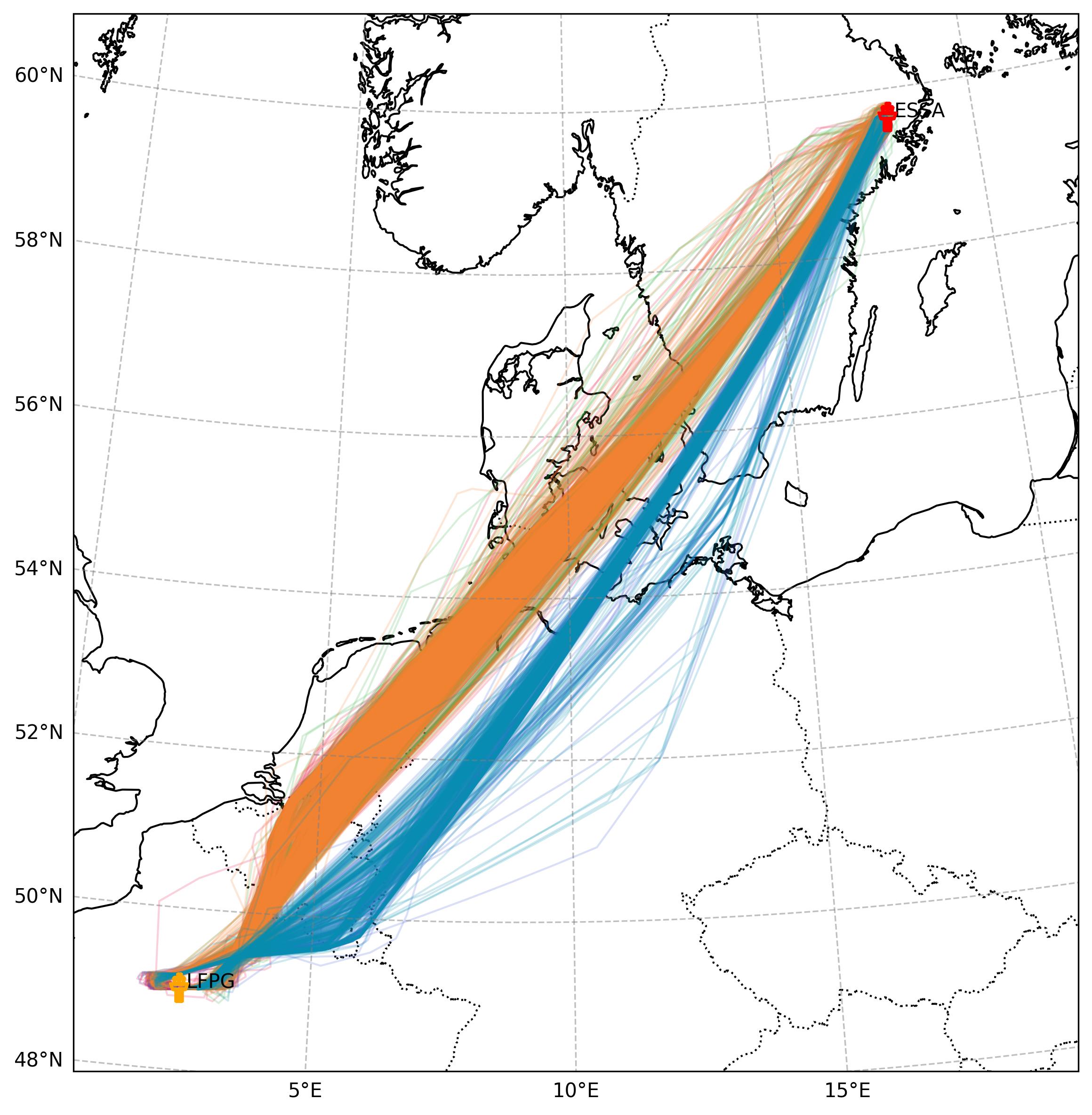}
        \caption{Real Trajectories}
    \end{subfigure}
    \hfill
    \begin{subfigure}{0.48\columnwidth}
        \includegraphics[width=\columnwidth]{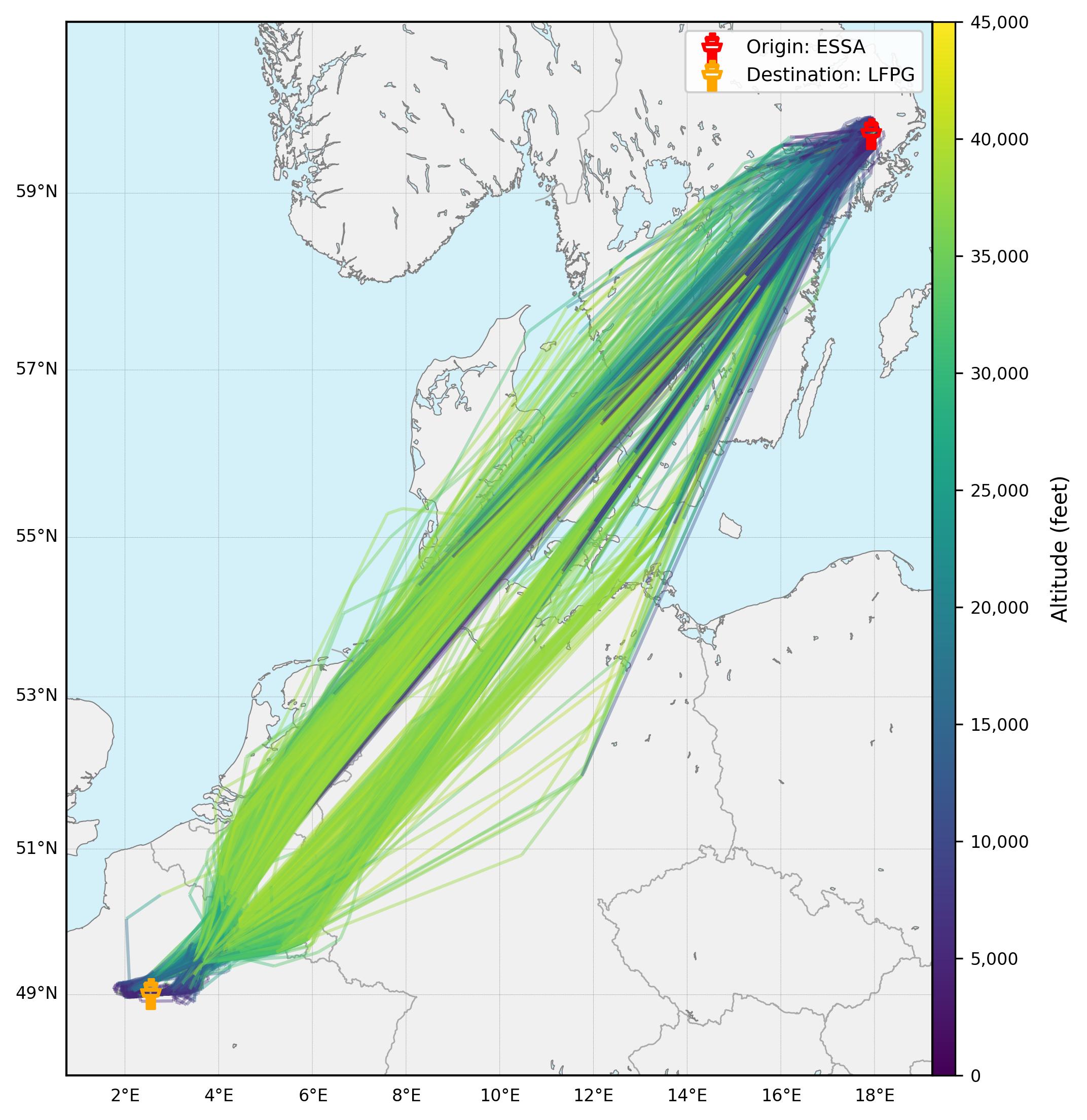}
        \caption{Real Altitude Profile}
    \end{subfigure}
    \vspace{0.2cm}

    \begin{subfigure}{0.48\columnwidth} 
        \includegraphics[width=\columnwidth]{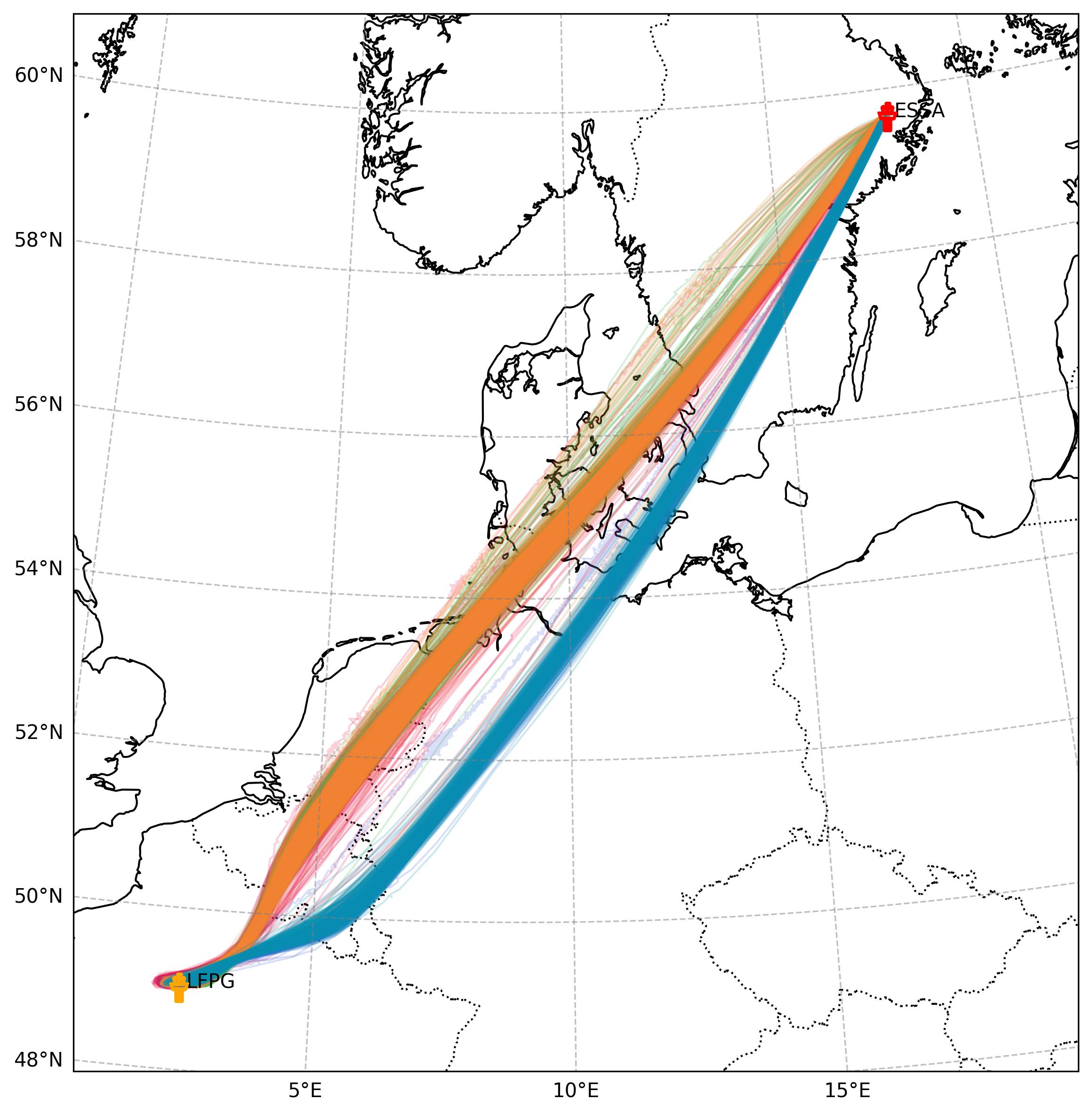}
        \caption{Synthetic Trajectories}
    \end{subfigure}
    \hfill
    \begin{subfigure}{0.48\columnwidth}
        \includegraphics[width=\columnwidth]{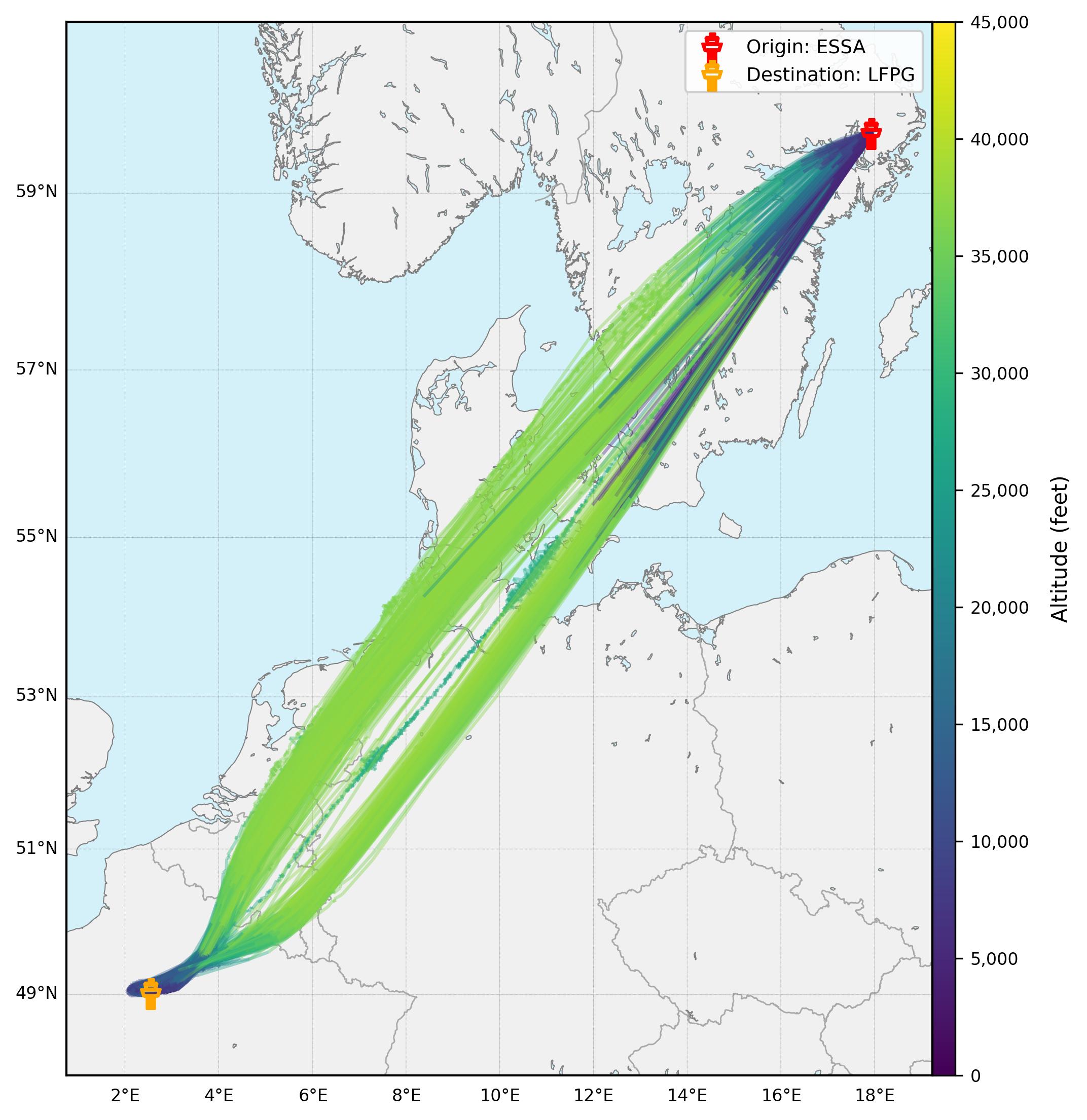}
        \caption{Synthetic Altitude Profile}
    \end{subfigure}
    \caption{Comparison of Real and Synthetic Trajectories (ESSA to LFPG) in Terms of Spatial Routes and Altitude Profiles.}
    \label{fig:trajectory_comparison_2}
\end{figure}

Figure \ref{fig:trajectory_comparison_2} provides further comparison between real and synthetic trajectories for the route Stockholm (ESSA) to Paris (LFPG). The generated samples capture general approach paths but omit some rare variations. Temperature tuning in the bidirectional transformers can increase diversity, but it may also yield spurious routes if set too high.

\subsection{Limitations and Future Work}

Although the model successfully captures several spatial and temporal features of real flight data, it occasionally produces trajectories with implausibly high speeds and underrepresents the broader variability seen in operational flights. These issues suggest a need for further refinement—particularly in implementing constraint-based or physics-informed learning to keep speeds and other parameters within realistic bounds. Future work will thus focus on integrating domain-specific constraints, refining the model’s capacity to capture outlier events, and enhancing its overall ability to represent the full scope of flight behaviors found in real-world data.

Additionally, our initial exploration of conditional generation centered on class labels (e.g., trajectory clusters). A natural extension is to incorporate weather-based conditional information, enabling the synthesis of flight paths adapted to varying atmospheric conditions. Doing so will require integrating external weather datasets and developing new validation metrics to quantify how weather variables influence the quality and realism of the generated trajectories.

\section{Conclusion}

This paper presents an adaptation of the Time-Based Vector Quantized Variational Autoencoder (TimeVQVAE) for generating synthetic aircraft trajectories. We leverage the model's architecture, which combines time-frequency domain processing, vector quantization, and transformer-based priors, to capture the complex spatiotemporal dependencies inherent in flight data. Empirical evaluations, including standard quality metrics, statistical tests, and a simulator-based flyability assessment, indicate that TimeVQVAE generally outperforms a temporal convolutional VAE baseline. Visual inspections corroborate these findings, demonstrating that the model successfully reproduces important flight characteristics observed in real data—albeit with some limitations in capturing the full range of operational variations. Overall, this work highlights both the utility of TimeVQVAE for data-driven trajectory modeling and opportunities for improvement through tighter domain constraints, weather conditioning, and enhanced variability modeling.

\section*{Acknowledgment}

This paper is based on research conducted within the SynthAIr project, which has received funding from the SESAR Joint Undertaking under the European Union’s Horizon Europe research and innovation program (grant agreement No. 101114847). The views and opinions expressed in this paper are solely those of the authors and do not necessarily reflect those of the European Union or the SESAR 3 Joint Undertaking. Neither the European Union nor the SESAR 3 Joint Undertaking can be held responsible for any use of the information contained herein.

We would also like to thank Daesoo Lee for his valuable help in implementing the TimeVQVAE model.

\balance

\bibliographystyle{IEEEtran}

\end{document}